\documentclass[british,format=sigconf]{aamas}
\usepackage[latin9]{inputenc}
\usepackage{color}
\usepackage{amsbsy}
\usepackage{amsthm}

\usepackage{graphicx}
\PassOptionsToPackage{normalem}{ulem}
\usepackage{ulem}

\makeatletter

\usepackage[linesnumbered,ruled]{algorithm2e}
\usepackage{url}
\usepackage{capt-of}
\usepackage{hhline}

\setcopyright{ifaamas}
\acmConference[AAMAS '22]{Proc.\@ of the 21st International Conference
on Autonomous Agents and Multiagent Systems (AAMAS 2022)}{May 9--13, 2022}
{Online}{P.~Faliszewski, V.~Mascardi, C.~Pelachaud,
M.E.~Taylor (eds.)}
\copyrightyear{2022}
\acmYear{2022}
\acmDOI{}
\acmPrice{}
\acmISBN{}

\author{Dung Nguyen}
\affiliation{
  \institution{A$^2$I$^2$, Deakin University}
  \city{Geelong}
  \country{Australia}}
\email{dung.nguyen@deakin.edu.au}

\author{Phuoc Nguyen}
\affiliation{
  \institution{A$^2$I$^2$, Deakin University}
  \city{Geelong}
  \country{Australia}}
\email{phuoc.nguyen@deakin.edu.au}

\author{Svetha Venkatesh}
\affiliation{
  \institution{A$^2$I$^2$, Deakin University}
  \city{Geelong}
  \country{Australia}}
\email{svetha.venkatesh@deakin.edu.au}

\author{Truyen Tran}
\affiliation{
  \institution{A$^2$I$^2$, Deakin University}
  \city{Geelong}
  \country{Australia}}
\email{truyen.tran@deakin.edu.au}

\acmSubmissionID{439}

\newcommand{\BibTeX}{\rm B\kern-.05em{\sc i\kern-.025em b}\kern-.08em\TeX}


\renewcommand{\citet}{\citep}
\usepackage{balance}
\usepackage{microtype}

\@ifundefined{showcaptionsetup}{}{%
 \PassOptionsToPackage{caption=false}{subfig}}
\usepackage{subfig}
\makeatother

\usepackage{babel}
\begin{document}
\pagestyle{fancy} 
\fancyhead{}
\title{Learning to Transfer Role Assignment Across Team Sizes}
\begin{abstract}
Multi-agent reinforcement learning holds the key for solving complex
tasks that demand the coordination of learning agents. However, strong
coordination often leads to expensive exploration over the exponentially
large state-action space. A powerful approach is to decompose team
works into roles, which are ideally assigned to agents with the relevant
skills. Training agents to adaptively choose and play emerging roles
in a team thus allows the team to scale to complex tasks and quickly
adapt to changing environments. These promises, however, have not
been fully realised by current role-based multi-agent reinforcement
learning methods as they assume either a pre-defined role structure
or a fixed team size. We propose a framework to learn role assignment
and transfer across team sizes. In particular, we train a role assignment
network for small teams by demonstration and transfer the network
to larger teams, which continue to learn through interaction with
the environment. We demonstrate that re-using the role-based credit
assignment structure can foster the learning process of larger reinforcement
learning teams to achieve tasks requiring different roles. Our proposal
outperforms competing techniques in enriched role-enforcing Prey-Predator
games and in new scenarios in the StarCraft II Micro-Management benchmark. 
\end{abstract}
\maketitle

\section{Introduction \label{sec:Introduction}}

\inputencoding{latin9}Learning to work as a team is essential to achieve
larger collective goals in solving complex tasks \citet{bucsoniu2010multi,zhang2021multi}.
However, partial observations and expensive team coordination may
prevent agents from having full knowledge of the environment and all
others operating on it \citet{omidshafiei2017deep,wang2020rode}.
Early work avoided this difficulty by learning independent single-agent
policies and treating other learning agents as part of the environment
\citet{matignon2012independent,tan1993multi}, but may run into the
non-stationary problem \citet{wang2020rode}. Centralised Training
Decentralised Execution (CTDE) \citet{rashid2018qmix,oliehoek2008optimal,kraemer2016multi}
is a middle ground assuming that the agents act on their own\emph{
after being trained together}. The essence of CTDE is to learn to
assign credit to individual agents when the whole team is trained
to maximise collective rewards. There has been a growing effort to
solve CTDE \citet{mahajan2019maven,rashid2018qmix,tonghan2020roma,lowe2017multi,foerster2018counterfactual,gupta2017cooperative},
but these are typically limited to small teams of agents since it
is prohibitive to explore the joint state-action space of large teams
during training. Learning with a large number of agents remains very
challenging \citet{samvelyan2019starcraft,wang2020rode}.

\begin{figure}[t]
\begin{centering}
\includegraphics[width=0.65\columnwidth]{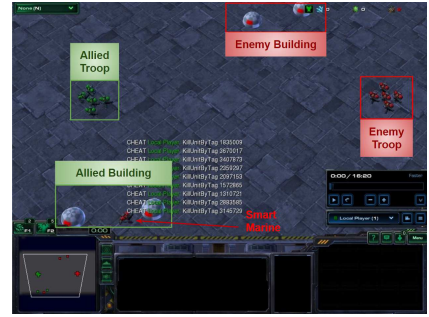} 
\par\end{centering}
\caption{\label{fig:SMACBuildingsSmartMarine} A StarCraft II scenario in which
the agents need to play different roles to explore good policies.
Some marines need to defend own campsites (bottom) under attack, enabling
their allies to attack the opponent's campsites (top).}
\end{figure}

A solution found to be effective in human teams is to decompose a
large team task into sub-tasks and roles. Under this decomposition,
each team member assumes one or more roles associated with manageable
sub-tasks \citet{biddle2013role}. Training a large team is therefore
feasible as each individual needs to explore only a constrained state-action
space defined by the assigned roles. However, in practice, the role
structures are not always well-defined \emph{a priori} or are changing
due to the task or team dynamics. In these circumstances, members
need to assume emerging roles and proactively play the chosen roles
when they see fit \citet{tonghan2020roma}. Learning to play emergent
roles essentially boils down to how to assign credits to roles followed
by assignments to the agents who play the roles \citet{wang2020rode}.

An orthogonal approach in solving difficult tasks is to \emph{learn
with a curriculum} in that we start from a small, easy-to-learn task,
then progressively expand the reach to larger tasks \citet{elman1993learning,narvekar2019learning,narvekar2020curriculum}.
For example, it would be learning from a small, simple environment
first, and gradually training in larger, more complex environments.
In multi-agent settings, it could be progressing from training a small
team where coordination is easy and cheap, then transferring the learned
skills to the next phase of training with a larger team where coordination
is difficult and expensive \citet{weixun2019few2more}. This curriculum
strategy demands a new kind of models that can work across team sizes.

In this work, we seek to bring these two approaches into a unified
learning framework for CTDE, which consists of (a) team learning to
assign credit to roles, and (b) transferring models (both credit assignment
and individual policy) across varying team sizes. We start from a
popular CTDE framework known as QMIX \citet{rashid2018qmix}, which
has a mixing network to aggregate $Q$-value functions of individual
agents into the team's $Q$-value function. The mixing network is
generated by a \emph{hyper-network} \citet{ha2016hypernetworks} that
takes as input the state, and thus assuming a fixed team size. Lifting
this constraint, we design a new mixing network and a new way to generate
the network from observations local to each agent rather than from
the entire state of the system. The design of the mixing network permits
(a) dynamic credit assignments to hidden roles, and (b) role assignments
to individual agents. At each time step, the network estimates the
probability that an agent will contribute to a role, collects the
$Q$-function values attributed to the role, and weights the role's
contribution to the total team's $Q$-function. Crucially, our \emph{generating
hyper-network is transferable across teams} by permitting varying
team sizes to borrow pre-trained models. This enables faster training
in a new setting and curriculum learning from easy to complex scenarios.

To further encourage the role differentiation and assignment among
agents, we introduce role-induced losses. For concreteness, we study
a \emph{loss associated with the reward horizons}, as encapsulated
in the discount factors in the MDP. This is motivated by the fact
that we humans engage in playing a long-term rewarded role for the
whole team even when we know the role has no short-term individual
benefit. We evaluate our proposed framework on two suites of multi-agent
reinforcement learning (MARL) experiments, highlighting the need for
curriculum learning when solving strongly cooperative CTDE tasks.
The first suite consists of enriched Prey-Predator games, where agents
must learn to recognise, pick and play an emergent role for the team
to succeed. Visual examination clearly shows that (a) the roles emerge,
and (b) curriculum learning enabled by transferring role assignment
is critical to success in larger teams. The other suite of experiments
is derived from the popular StarCraft Multi-agent challenge (SMAC)
\citet{samvelyan19smac} where we enforce stricter team coordination.
Again we demonstrate that ours converges much faster than competing
methods when tested on larger teams, thanks to the ability to transfer
role assignment from smaller teams.

To summarise, our contributions are: (1) A neural framework that learns
to assign credits and agents to roles, and \emph{supports transferring
across different team sizes}. The latter further enables curriculum
learning to succeed in larger team settings and more complex tasks;
and (2) An enriched version of Prey-Predator games and a suite of
new scenarios in SMAC to support varying reward horizons.

\section{Preliminaries}

\paragraph{Cooperative Reinforcement Learning Agents}

\inputencoding{latin9}We consider a \emph{fully cooperative multi-agent
task }described as a decentralised Partial Observability Markov Decision
Process (Dec-POMDP) \citet{oliehoek2016concise} $G=\left\langle \mathcal{N},\mathcal{S},\mathcal{A},\mathcal{P},R,\Omega,\mathcal{O},\gamma\right\rangle $
in which $\mathcal{N}$ is the set of agents, $s\in\mathcal{S}$ is
the true state of the environment, $\mathcal{O}$ is the observation
space, $\mathcal{A}$ is the action space, $\Omega:\mathcal{S}\longmapsto\mathcal{O}$
is a mapping from state space to the observation space, $\mathcal{P}:\mathcal{S}\times\mathcal{A}^{|\mathcal{N}|}\times\mathcal{S}\longmapsto\left[0,1\right]$
is the state transition function, $R:\mathcal{S}\times\mathcal{A}^{|\mathcal{N}|}\longmapsto\mathbb{R}$
is the reward function, and $\gamma\in\left[0,1\right]$ is the discounted
factor. At each time step $t$, an agent $i$ observes its local observation
$o_{i}^{(t)}\in\mathcal{O}$ then decides its action $a_{i}^{(t)}\in\mathcal{A}$.
All agents in the environment form a joint-action $a\in\mathcal{A}^{|\mathcal{N}|}$.

\paragraph{Centralised Training and Decentralised Execution (CTDE)}

In this setting, we train a team of agents together to maximise team
rewards resulting from individuals executing their own policies $\pi^{i}:\mathcal{O}\longmapsto\mathcal{A}$
with $i\in\mathcal{N}$. Once trained, agents make decisions based
on their own local observations. The CTDE scheme allows the team of
agents to know the true state of the environment $s\in\mathcal{S}$
and all observations of others $o_{j}\in\mathcal{O}$ for all $j\in\mathcal{N}$
during training. Howerver, an agent can only access to its local observations
$o_{i}^{(t)}$ to make decision $a_{i}^{(t)}$ during execution to
maximise the team reward ${\displaystyle R_{\text{team}}=\sum_{i\in\mathcal{N}}\sum_{t=0}^{T}\gamma^{t}r_{i}^{(t)}}$.

\section{Proposed Method}

\inputencoding{latin9}\begin{figure}
\begin{centering}
\includegraphics[width=0.9\columnwidth]{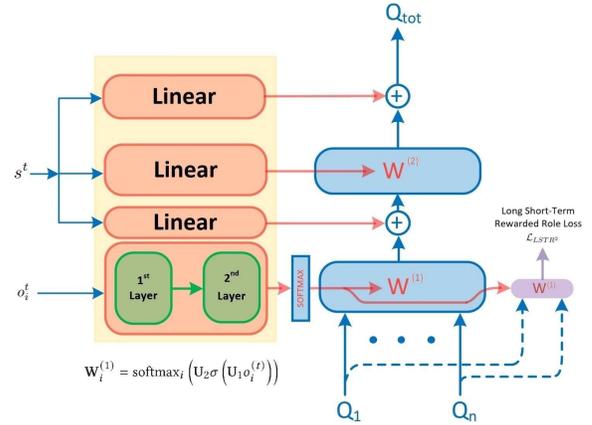} 
\par\end{centering}
\caption{\label{fig:OurCDTEModel} Our architecture to train a team of reinforcement
learning agents in Centralised Training Decentralised Execution (CTDE)
setting. Using the observations of agents in the first hyper-network
enables the ability to transfer the pre-trained model across team
sizes. There is no gradient passing through the dash lines. }
\end{figure}

Under the CTDE scheme, our aim is to design an architecture that learns
to factorise the total value function of the team $Q^{tot}(s,\boldsymbol{a})$
into $N=\vert\mathcal{N}\vert$ components, i.e. each agent $i$ will
predict a value $Q_{i}(o_{i},a_{i})$ based on its local observation
$o_{i}$ and action $a_{i}$. That is, $Q^{tot}$ is a mixing function,
computed by the\emph{ mixing network} whose input is the \emph{variable-size}
set of mixed values $\mathbf{Q}=\left\{ Q_{1},Q_{2},...,Q_{N}\right\} $.
The challenge is in learning the mixing network to properly assign
credits to individual agents who play the emerging roles as the team
interacts with the environment. The overall network design is given
in Fig.~\ref{fig:OurCDTEModel}.

Each value function $Q_{i}$ is computed using a recurrent neural
network that takes the current observation, its own execution trajectory,
and a possible action. Given the local observation $o_{i}^{(t)}$
at time step $t$, the agent $i$ computes its prediction of action-value
$Q_{i}(o_{i}^{(t)},a_{i}^{(t)})=\text{MLP}\left(h_{i}^{(t)}\right)$
where $x_{i}^{(t)}=\text{ReLU}\left(\text{MLP}(o_{i}^{(t)})\right)$
is a transformation of the observation and $h_{i}^{(t)}=\text{GRU}\left(x_{i}^{(t)},h_{i}^{(t-1)}\right)$
is the hidden states of the gated recurrent neural networks (GRU)
\citet{chung2014empirical}.

\subsection{The Mixing Network \label{subsec:The-Mixing-Network}}

Taking the role-based approach and assuming there are $K$ ``roles'',
we design a new neural architecture for the mixing network. Given
the prediction of individual agents $\left\{ Q_{i}\right\} _{i=1}^{N}$,
the mixing network makes a prediction about the team reward: 
\begin{align}
Q^{tot} & =b^{(2)}+\sum_{k=1}^{K}\mathbf{W}_{k}^{(2)}\sigma\left(\mathbf{b}_{k}^{(1)}+\sum_{i=1}^{N}\mathbf{W}_{ik}^{(1)}Q_{i}\right),\,\text{s.t.}\label{eq:mixing-func}\\
\mathbf{W}_{ik}^{(1)} & \ge0;\,\,\sum_{i}\mathbf{W}_{ik}^{(1)}=1;\,\,\text{and}\,\mathbf{W}_{k}^{(2)}\ge0\,\,\text{for all}\,i,k,
\end{align}
where $\left\{ \mathbf{W}^{(1)},\mathbf{W}^{(2)}\right\} $ are mixing
coefficients, $\left\{ \mathbf{b}^{(1)},b^{(2)}\right\} $ are biases,
and $\sigma(\cdot)$ is an activation function chosen to be Exponential
Linear Unit in our implementation.

The mixing coefficient $\mathbf{W}_{ik}^{(1)}$ measures the contribution
of each agent $i$ to a role $k$. The normalisation over agents $\sum_{i}\mathbf{W}_{ik}^{(1)}=1$
for all $k$ can be interpreted as the \emph{probabilities we use
to select the agents} for each role. The mixing coefficient $\mathbf{W}_{k}^{(2)}$
assigns the credit to a role $k$ in the total estimated reward.

The mixing function in Eq\@.~(\ref{eq:mixing-func}) was first studied
in QMIX \citet{rashid2018qmix} in that the mixing coefficients and
biases are state-dependent, i.e., through hyper-networks (e.g., see
\citet{ha2016hypernetworks}) or fast weight (e.g., see \citet{hinton1987using,schmidhuber1992learning}).
However, as QMIX uses the global information (state) to compute $\mathbf{W}^{(1)}$,
it must assume a fixed team size without role assignment, and thus
cannot transfer the mixing network across different team sizes with
different roles.

To tackle this drawback, we design the hyper-network of the first
layer such as it receives the local observations of the agents as
inputs instead of the global states. The hyper-network generates the
mixing coefficients as follows:

\[
\mathbf{W}_{i}^{(1)}=\text{softmax}_{i}\left(\mathbf{U}_{2}\sigma\left(\mathbf{U}_{1}o_{i}^{(t)}\right)\right),
\]
where $o_{i}^{(t)}\in\mathbb{R}^{D}$ is the observation vector of
the agent $i$, $\mathbf{U}_{1}\in\mathbb{R}^{H\times D}$ and $\mathbf{U}_{2}\in\mathbb{R}^{K\times H}$
are the weights. Other mixing coefficients and biases are computed
similarly to those in QMIX: $\mathbf{b}_{1}^{(1)}=\text{MLP}\left(s^{(t)}\right)$;
$b^{(2)}=\text{MLP}\left(\sigma\left(\text{MLP}\left(s^{(t)}\right)\right)\right)$
and $\mathbf{W}^{(2)}=\text{MLP}\left(\sigma\left(\text{MLP}\left(s^{(t)}\right)\right)\right)$.

\paragraph{Remark}

Our architecture enables the ability to \emph{transfer the mixing
networks across different team sizes}. Therefore, this helps train
a team to solve difficult tasks with a smaller number of samples,
even in situations that could not be solved by QMIX.

\subsection{Role-Specific Reward Horizons \label{subsec:RoleBasedDecomposition}}

The effective time horizon for an action is often encapsulated in
the discount factor $\gamma$ of the expected future rewards. However,
specifying the discount factor remains an art. We hypothesise that,
in general, roles are best played with a given time horizon: some
roles are biased towards immediate rewards (e.g., shooting prey in
sight), while others are gearing towards the long-term (e.g., guarding
the camp).

This suggests the following regulariser at each time step $t$ of
a training episode:

\begin{equation}
\mathcal{R}_{LSTRR}^{(t)}=\frac{1}{K_{2}}\sum_{k=1}^{K_{2}}\left(Q_{k}^{*}-R_{k}\right)^{2}\,\,\text{for\,}\,K_{2}\le K,\label{eq:LSTRR}
\end{equation}
where LSTRR stands for \emph{Long-Short Term Rewarded Roles}. Here
$Q_{k}^{*}=\sigma\left(\sum_{i\in\mathcal{N}}\mathbf{W}_{ik}^{(1)}Q_{i}+b_{k}^{(1)}\right)$
is an estimation of the $Q$-value associated with role $k$, and
$R_{k}=\sum_{\tau=0}^{T-t}\gamma_{k}^{\tau}r^{(t+\tau)}$ is the discounted
reward for role $k$. This regulariser is used during the centralised
training process while we know the rewards. Without loss of generality
we assume $\gamma_{1},\gamma_{2},...,\gamma_{K_{2}}$ is a decreasing
sequence (from long-term to short-term horizons). In practice, we
choose $K_{2}=\left\lceil \frac{K}{2}\right\rceil $ to compute the
summation of all ${\displaystyle Q_{k}^{*}}$ ($k\in\left[1,K_{2}\right],k\in\mathbb{N}$)
before concatenating to $K-K_{2}$ components and multiplying with
$\mathbf{W}_{k}^{(2)}$ for $k\in\left[1,K-K_{2}+1\right],k\in\mathbb{N}$.

\subsection{Scaling Team Sizes by Curriculum Learning \label{subsec:Scaling-Team-Sizes}}

It has been observed that training a large team in CTDE is difficult
\citet{samvelyan2019starcraft,wang2020rode}. Thus we propose curriculum-based
learning. We start by training a small team then transfer to larger
teams, thus effectively reusing learnt models. Transferring across
team sizes is possible thanks to the design of the mixer which takes
as input agent-specific observations instead of the full observation
of the whole team.

Training a smaller team permits learning by demonstrations from experts.
Thus, it suggests a two-phase training procedure: (i) pre-train a
small team on experiences from experts using a supervised loss $\mathcal{L}_{sup}^{(t)}$,
and (ii) continue to train on a larger team through interacting with
the environment using a the temporal difference (TD) loss $\mathcal{L}_{TD}^{(t)}$.
Both steps can be expressed in the following joint loss function:
\begin{equation}
\mathcal{L}^{(t)}=\mathcal{L}_{TD}^{(t)}+\lambda_{1}\mathcal{L}_{sup}^{(t)}+\lambda_{2}\mathcal{R}_{LSTRR}^{(t)},
\end{equation}
where $\lambda_{1}>0$ is the contributing factor of the demonstration
when possible, and $\lambda_{2}>0$ is the contributing factor of
the horizon regularisation defined in Eq.~(\ref{eq:LSTRR}) when
the reward horizon matter. The losses are defined as: 
\begin{align*}
\mathcal{L}_{sup}^{(t)} & =-\frac{1}{N}\sum_{i\in\mathcal{N}}\log p(a_{i}=\hat{a}_{i}|o_{i}^{(t)}),\\
\mathcal{L}_{TD}^{(t)} & =\left(Q^{tot}-\hat{Q}^{tot}\right)^{2}
\end{align*}
with $\hat{a}_{i}$ is the action in the demonstration and $\hat{Q}^{tot}$
is the expectation of the ground truth team rewards. During the second
phase, only team rewards are provided, so only the TD loss is used,
i.e., we set $\lambda_{1}=0$.

\section{Experiment Results \label{sec:Exp}}

\inputencoding{latin9}We validate our proposed method on two multi-agent
settings: An enriched version of the Prey-Predator game (Section~\ref{subsec:Enriched-Prey-Predator-Games})
and the popular StarCraft Multi-agent Challenge (SMAC) \citet{samvelyan19smac}
(Section~\ref{subsec:StarCraft-Multi-Agent-Challenge}).

\subsection{Enriched Prey-Predator Games \label{subsec:Enriched-Prey-Predator-Games}}
\begin{center}
\begin{figure}
\begin{centering}
\includegraphics[width=1\columnwidth]{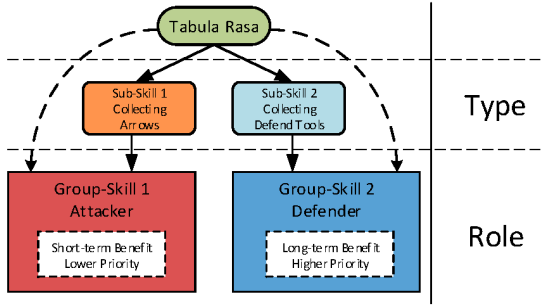} 
\par\end{centering}
\caption{\label{figModifiedStagHuntStructure} The structure of our enriched
prey-predator games. All agents are \emph{tabula rasa} since the beginning
of one episode. Each agent can always choose their type by collecting
the arrows (then be able to attack all prey on their sight without
being removed from the game) or collecting the defence tool (to be
able to enter the campsite). While attacking prey immediately brings
reward, defending campsites allows the team to explore more attack
strategies to obtain higher long-term rewards. }
\end{figure}
\par\end{center}

Prey-Predator is a canonical framework to study cooperative behaviours
in multi-agent learning. The original format has a team of predators
who may cooperate to catch prey. There is one near-optimal behaviour
at which all agents cooperate to catch the prey, e.g. only one type
of role is required.

\paragraph{Game enrichment}

We enriched the Prey-Predator framework to induce the concept of role
and to enforce strong coordination to achieve the task. In the new
game format, there are three \emph{types} of predators: (1) \emph{Normal},
(2) \emph{Archer}, and (3) \emph{Defender}. Each predator can choose
actions in the set of \{\texttt{\small{}Left, Right, Up, Down, Stay,
Catch, Skill-act}\}. After collecting \emph{arrows}, an archer can
use \texttt{\small{}Skill-act} to have a higher range of attacking
the prey, i.e., attacking all prey in its sight instead of only the
prey next to it. If an agent (normal, archer or defender) successfully
kills a prey by \texttt{\small{}Catch}, the agent will be removed
from the map \citet{bohmer2020deep}. However, if an archer uses\texttt{\small{}
Skill-act} to kill prey, they can continue to hunt. The next modification
is that we introduce camps on the map. If a prey steps into a camp,
the game is over with a reward of $-5.0$ for the predator team. An
agent chooses between at least two \emph{roles}: camp defender or
prey attacker. To defend the camps, the agent needs to collect a \emph{defence
tool}, steps into the campsite and stays there to prevent the prey
from jumping in. The structure of our enriched prey-predator games
is shown in Fig.~\ref{figModifiedStagHuntStructure}. 

Agents playing this enriched Prey-Predator can have suboptimal behaviours
and near-optimal behaviours. For example, one suboptimal behaviour
is when all agents (including the archer) try to catch prey without
collecting arrows. If the number of agents is insufficient for catching
all prey, the team that follows this joint policy will obtain low
rewards. In the best case, the team will kill all prey and keep campsites
clear. One effective strategy is to separate the team into two parts:
(1) some agents collect defence tools and stay inside the camps (they
are allowed to move out of the camps); and (2) other agents collect
arrows and kill all prey. Although this strategy seems obvious to
humans, it is a challenge for a team of artificial agents to learn.
\begin{center}
\begin{figure}
\begin{centering}
\includegraphics[width=0.25\columnwidth]{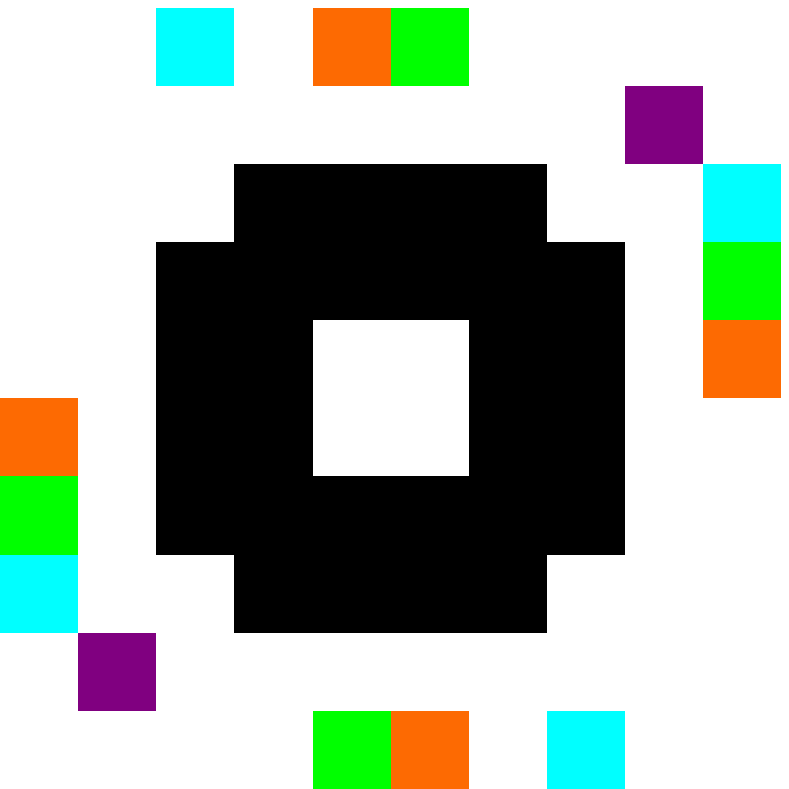}
\hspace{2em} \includegraphics[width=0.25\columnwidth]{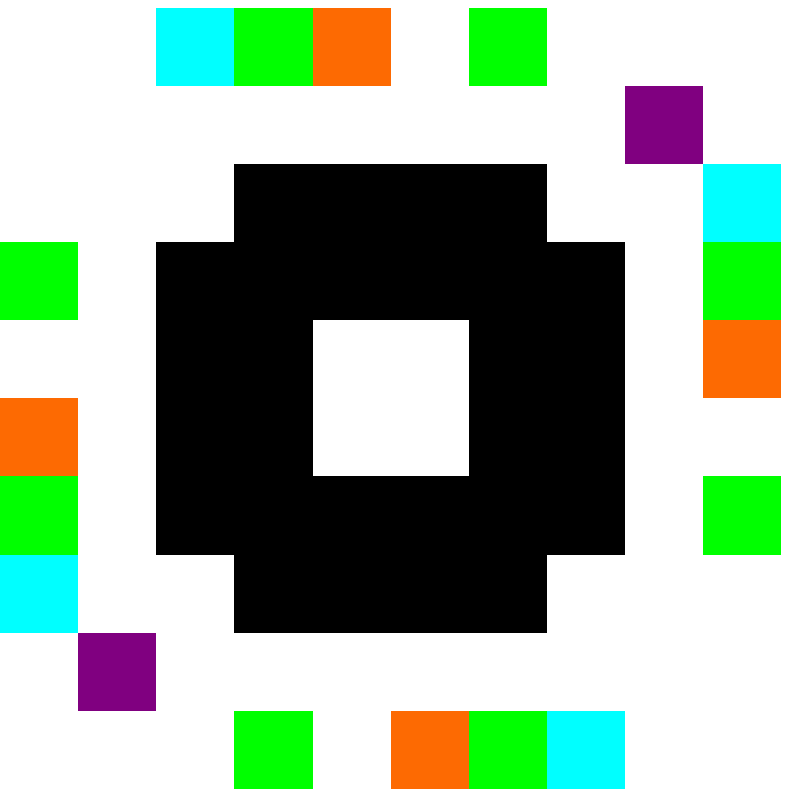}\hspace{2em}
\includegraphics[width=0.25\columnwidth]{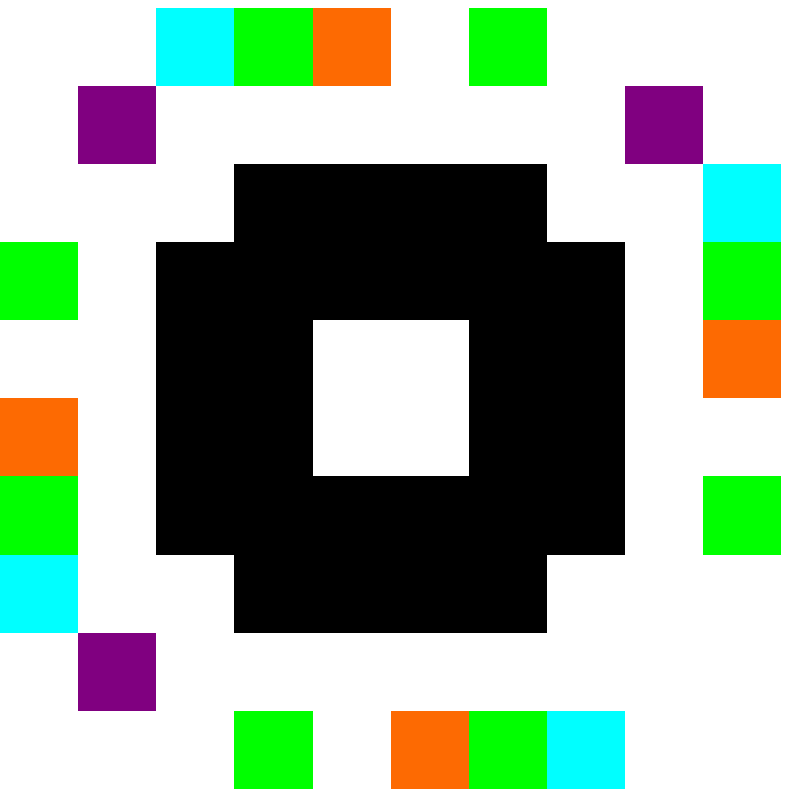}
\par\end{centering}
\caption{\label{figTrainingMap} Example maps of environments: \textbf{\uline{Left}}:
The map of \emph{pre-train} environment with 4 agents, 2 campsites;
\textbf{\uline{Middle}}: \emph{target} environment with 8 agents,
2 campsites; \textbf{\uline{Right}}: \emph{target} environment
with 8 agents, 3 campsites. In each map, there are five objects with
different colours: the prey (black); the agents (\textcolor{green}{green});
the arrows (\textcolor{orange}{orange}); (4) the defend tools (\textcolor{cyan}{cyan});
and the campsites (\textcolor{purple}{purple}).}
\end{figure}
\par\end{center}

To test the ability to play the defence role of agents, we set up
a \textit{smart prey} that will directly move toward the campsites
-- if it successfully gets there, the game will be over. The \textit{smart
prey} will reach the top right campsite after $3$ steps, which means
the top right agents need to strictly collect the defence tool and
jump into this campsite to defend. This will prevent the situation
that agents can find aggressive behaviour in attacking, i.e. directly
collect arrows and kill all potential prey before it jumps into the
campsites.

\paragraph{Transfer learning strategy}

To test transfer learning capability, we created environments of different
difficulties. Fig.~\ref{figTrainingMap} shows an example of easy
setting (4 agents, 2 campsites, 28 prey), moderate setting (8 agents
and 0-2 campsites) and difficult setting (8 agents, 3 campsites).
The model is first pre-trained on the easy environment using $50$
demonstrations. The trained team succeeds in defending all campsites
and capturing all the prey. Upon convergence, we continue to train
the model in the target environment by temporal differencing. 

\subsubsection{Transferring Results Across Team Sizes}

\begin{figure*}
\begin{centering}
\subfloat[Test maps of moderate settings. \label{fig:Test-maps-of-moderate}]{\begin{centering}
\includegraphics[height=1\columnwidth]{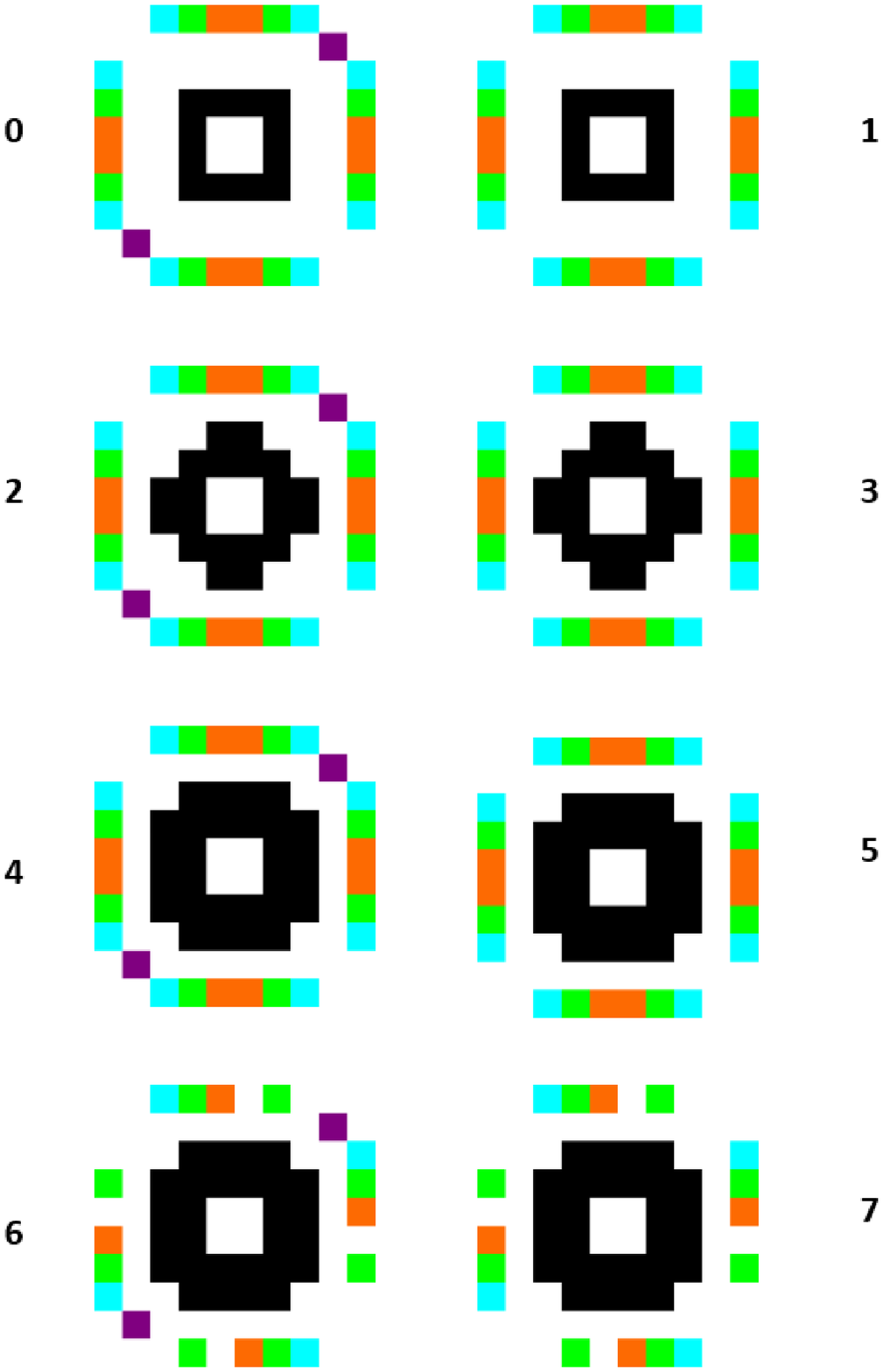}
\par\end{centering}
}\subfloat[Performance of transfer.\label{fig:Performance-of-transfer}]{\begin{centering}
\includegraphics[height=1\columnwidth]{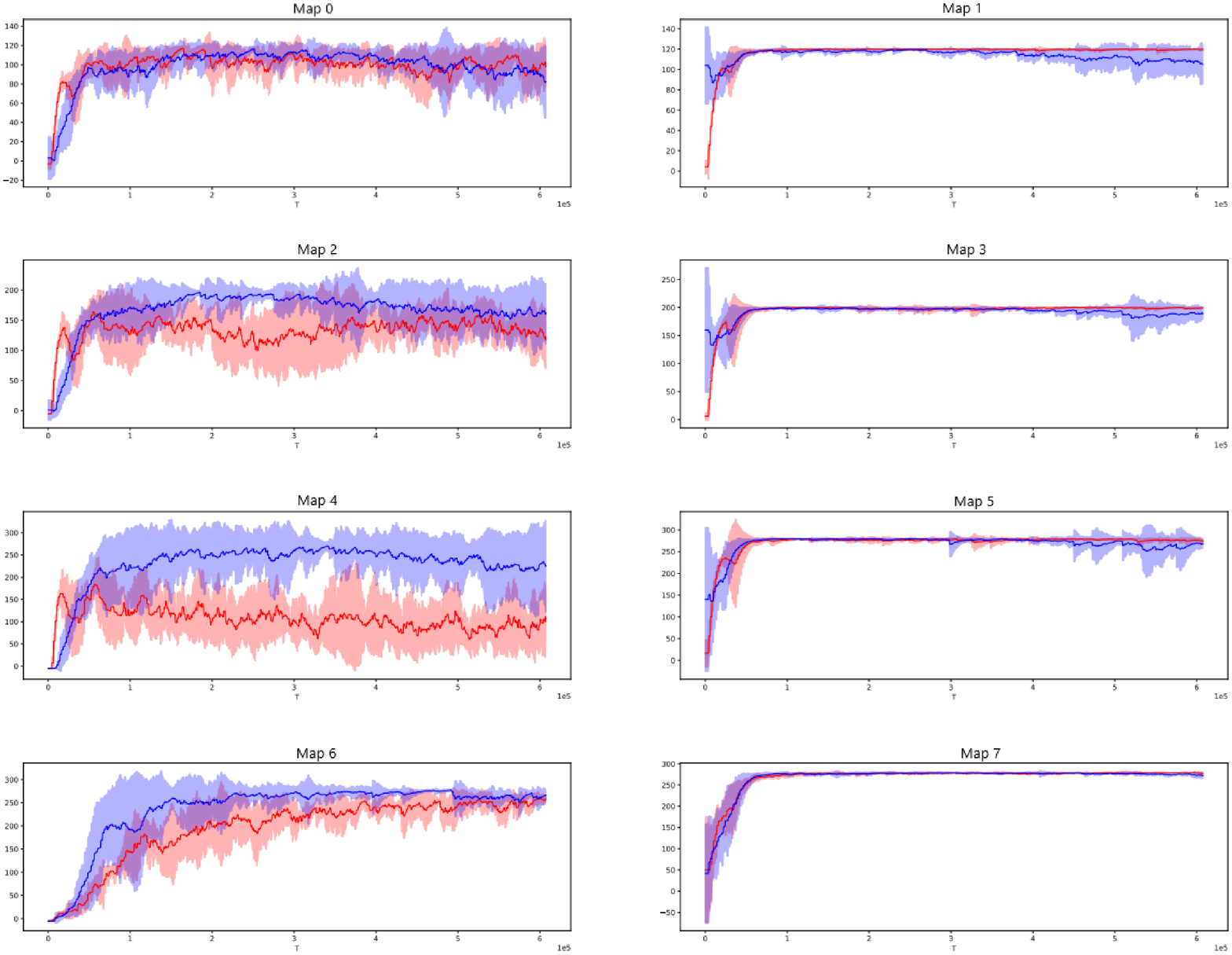}
\par\end{centering}
}
\par\end{centering}
\caption{\label{fig:TransferEasySetting} Moderate settings of enriched prey-predator
with 8 agents and 2 campsites. (a) Test maps in moderate settings
of modified prey-predator games with 8 agents. The maps with even
index has two campsites; the maps with odd index does not have any
campsite. (b) Team rewards (y-axis) vs. training time steps (x-axis)
of teams trained by QMIX (\textcolor{red}{red}) and our method (\textcolor{blue}{blue}).
Our method can learn the optimal behaviour faster than QMIX in the
training environment (map No.6). In map No.4, the team trained by
QMIX could not behave optimally, leading to a significant difference
in the team rewards.}
\end{figure*}

We first created 8 scenarios of moderate difficulty; each has 8 agents,
with or without campsites. Fig.~\ref{fig:Test-maps-of-moderate}
shows 8 maps. The corresponding performance curves are plotted in
Fig\@.~\ref{fig:Performance-of-transfer}. It can be seen from Fig\@.~\ref{fig:Performance-of-transfer}
that the agents in our team can learn the optimal behaviour faster
than the team trained by QMIX \citet{rashid2018qmix}. Furthermore,
in environments with different object positions (defence tools and
arrows) or a different amount of prey, our team can find the optimal
behaviour while the QMIX can not. 

\begin{flushleft}
\begin{figure}
\begin{centering}
\includegraphics[width=1\columnwidth]{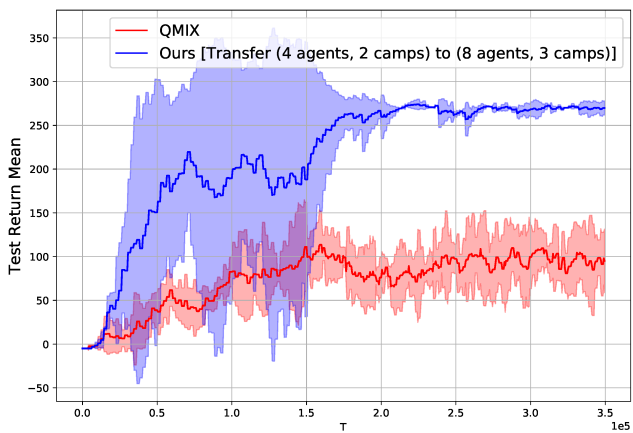}
\par\end{centering}
\caption{Team rewards (y-axis) vs. training time steps (x-axis) of teams trained
by QMIX (\textcolor{red}{red}) and our method (\textcolor{blue}{blue})
in the difficult setting which has 8 agents and 3 campsites. The team
trained by our method can learn to converge to optimal behaviour,
while the team trained by QMIX can not. \label{fig:Team-Rewards-vs-Time}}
\vspace{-6mm}
\end{figure}
\par\end{flushleft}

\begin{figure*}
\begin{centering}
\includegraphics[width=0.85\columnwidth]{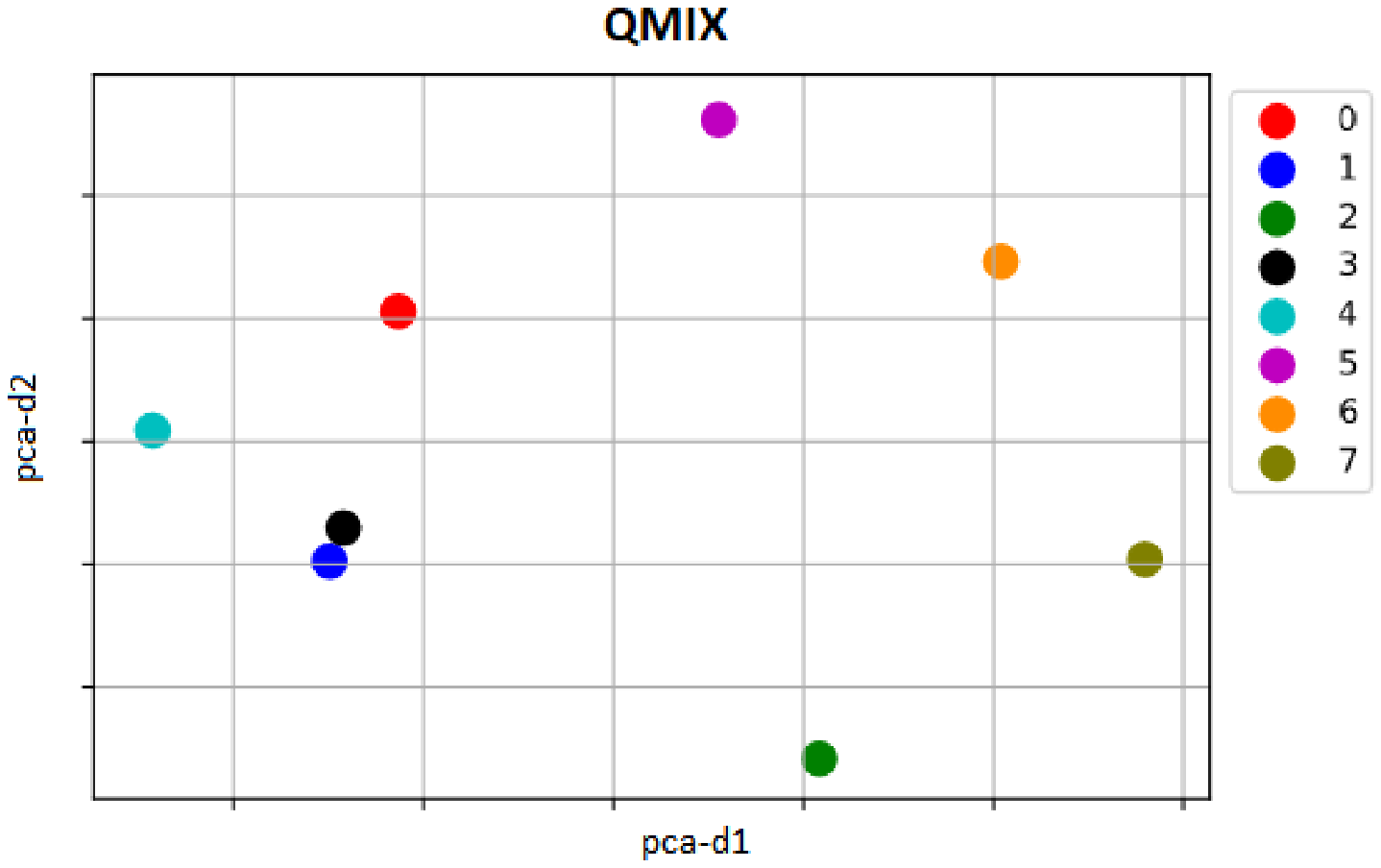}
\hspace{0.5cm}\includegraphics[width=0.85\columnwidth]{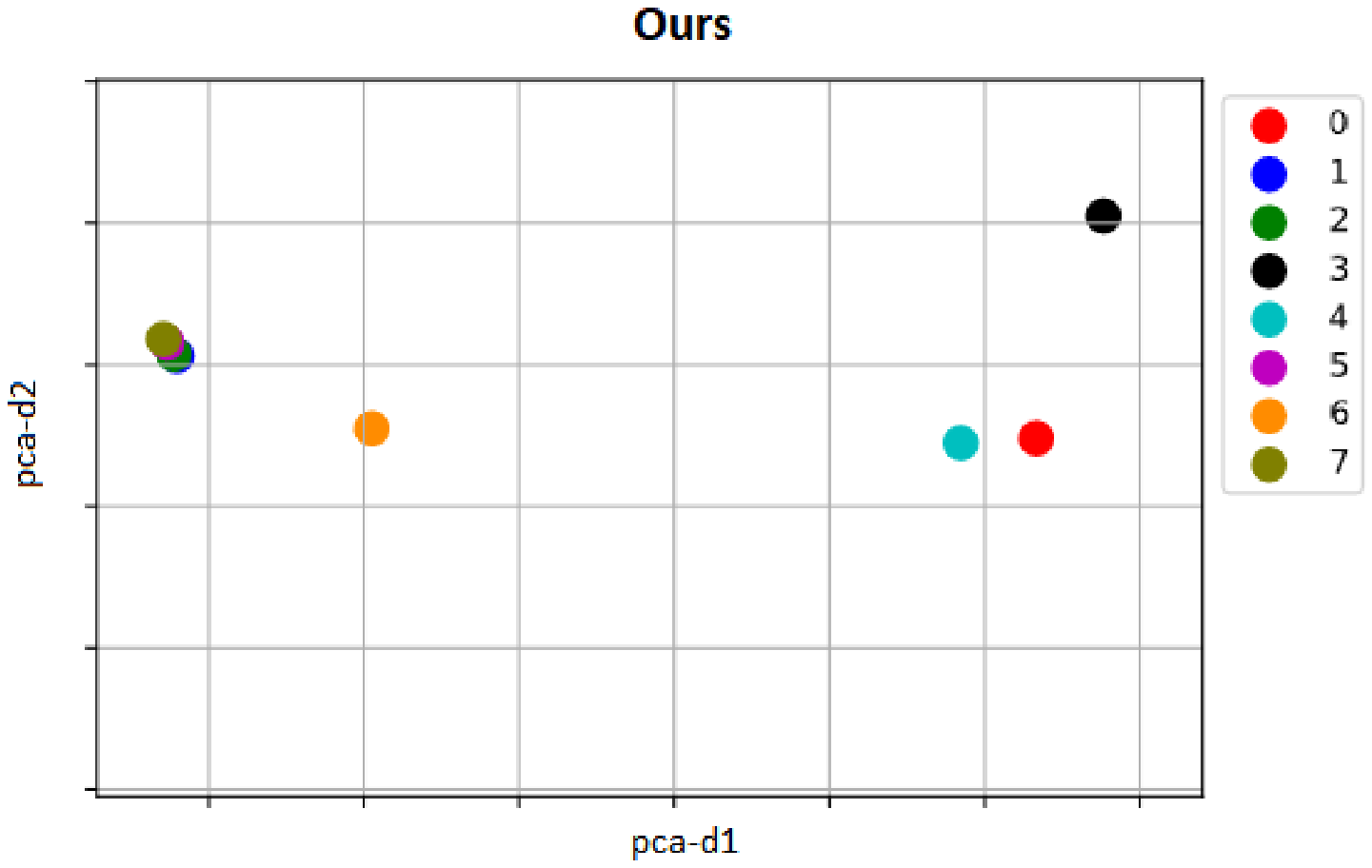}
\par\end{centering}
\caption{\label{fig:PCAMixerLatent} PCA projection of the first layer weight
$\mathbf{W}^{(1)}$ into 2D -- QMIX (left) and our architecture (right).
The number is the indices of agents. The group of agents $\{0,3,4\}$
which should collect arrows to capture prey is separated from the
group of agents $\{1,2,5,6,7\}$ which should defend the campsites.
We observe that amongst the set of defenders $\{0,3,4\}$, the agent
No. $3$, which is nearby the smart prey and should strictly defend
the campsite, has the $2\text{D}$ latent variable far from others
in the same group of agents playing defend role. Our method can learn
the optimal behaviour in the training environment, while the QMIX
could not learn the optimal policy.}
\end{figure*}

\begin{figure*}
\begin{centering}
\subfloat[\label{figVisitationMapQMIX} QMIX]{\begin{centering}
\includegraphics[width=0.9\columnwidth]{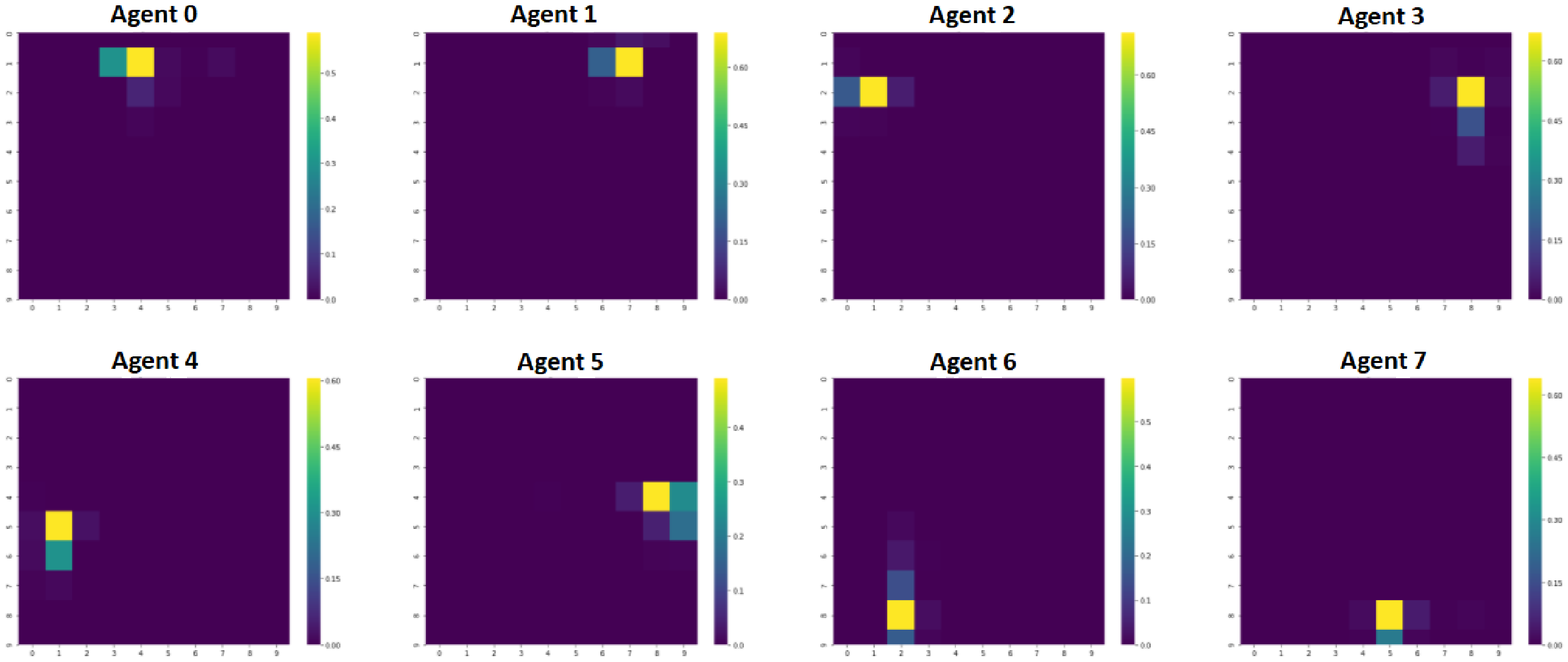}
\par\end{centering}
}\hspace{1.2cm}\subfloat[\label{figVisitationMapTransfer} Our method]{\begin{centering}
\includegraphics[width=0.9\columnwidth]{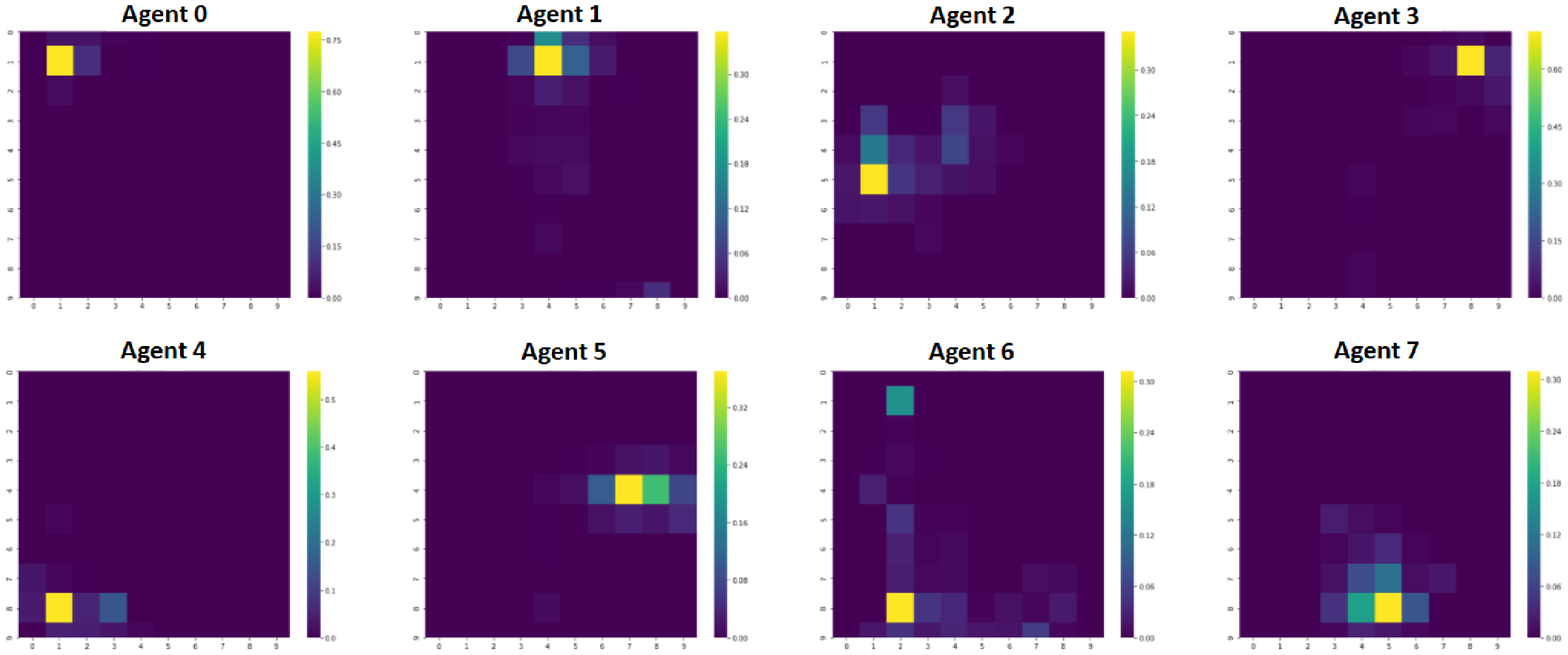}
\par\end{centering}
}
\par\end{centering}
\caption{\label{figVisitationMap}The visitation map at around $200,000$ time
steps of agents trained by (a) QMIX; and (b) Our method. There are
eight agents 0-based indexed. Their visitation map from left to right
and from top to down, for example, the first map in the second row
of each sub-figure is the visitation map of agent $4$. In the team
trained by QMIX, the agents could not learn to defend the campsites.
If trained by our method, agents indexed by ${0,3,4}$ should defend
the campsite. This does not only help to avoid failure but also enable
others to find out the optimal policy for capturing all prey (other
agents ${1,2,5,6,7}$ can explore the map, which is shown in sub-figure
b).}
\end{figure*}
We then make the target environment more challenging with 8 agents
and 3 campsites (e.g., see Fig.~\ref{figTrainingMap}(rightmost)).
Fig.~\ref{fig:Team-Rewards-vs-Time} shows the performance curves
of our architecture compared to QMIX on 8 agents and 3 campsites.
While the team of agents trained by QMIX gets stuck at sub-optimal
policies, our model can learn the optimal policy in which our team
obtain higher rewards by first defending all the campsites then capturing
all prey in the game. 

\subsubsection{Emerging Roles}

To understand the behaviour of the trained team, we project the mixing
coefficient which is generated by the first-layer hyper-network ($\mathbf{W}^{(1)}$
in Eq.~(\ref{eq:mixing-func})) for each agent onto 2D by PCA. Fig.~\ref{fig:PCAMixerLatent}
shows the difference between agents within different roles. The group
of agents $\{0,3,4\}$ which should collect arrows to capture prey
is separated from the group of agents $\{1,2,5,6,7\}$ which should
defend the campsites. Interestingly, the agent No. $3$, which is
nearby the smart prey and should strictly defend the campsite, has
the $2\text{D}$ latent variable far from others in the same group
of agents playing defend role. Agents $0,3$, and $4$ are placed
nearby the defence tools and the campsites. Therefore, they have higher
frequencies of visiting the campsites to defend, while other agents
learn to collect the arrows and move around together to capture all
prey. The visitation map is shown in Fig.~\ref{figVisitationMap}.

\subsubsection{Ablation Study}

\begin{figure}[thp]
\begin{centering}
\includegraphics[width=0.85\columnwidth]{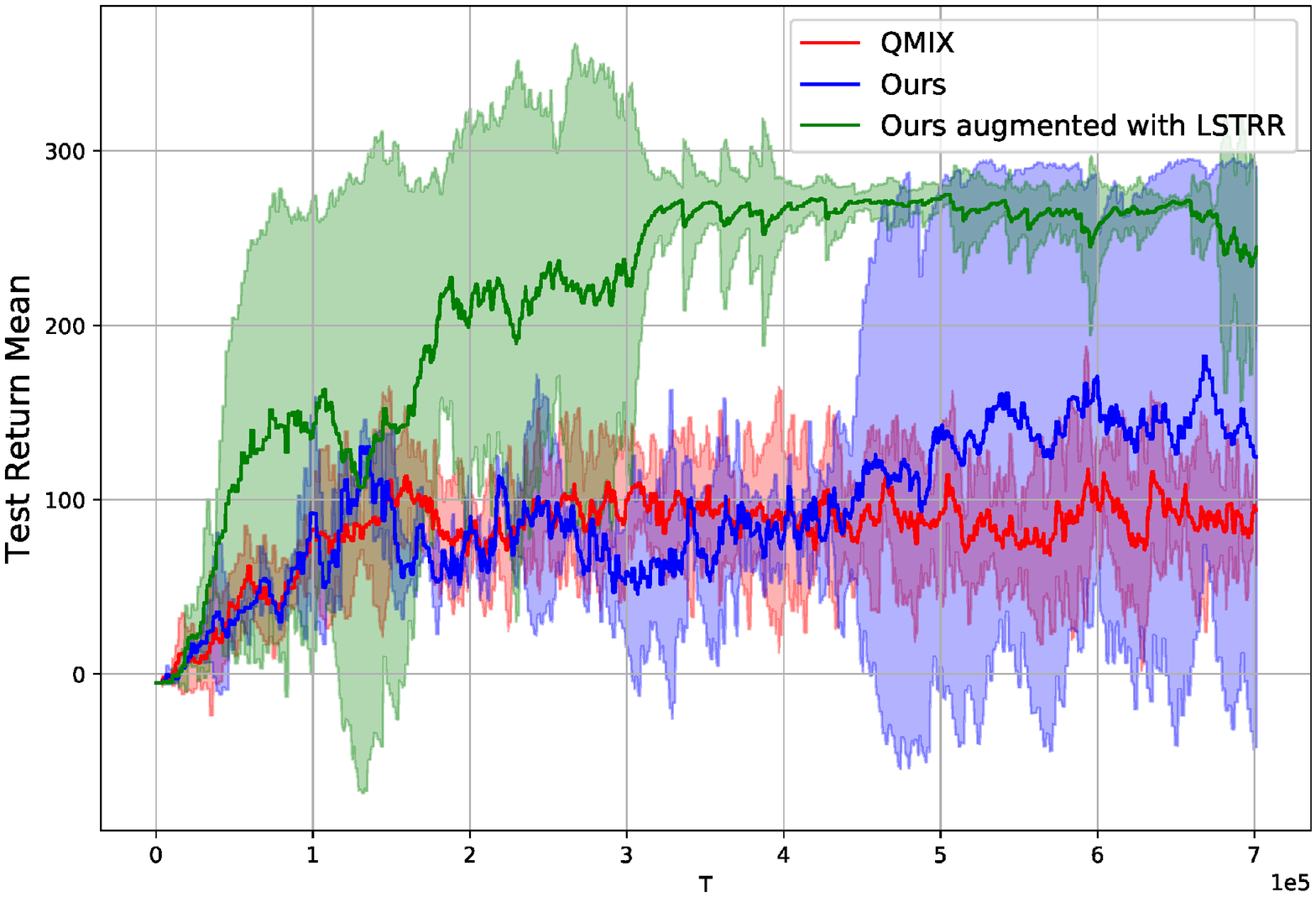}
\par\end{centering}
\caption{\label{fig:figTransferHardSettingLSTR} Team rewards (y-axis) vs.
Training time steps (x-axis) of teams trained by QMIX (\textcolor{red}{red})
and our method (\textcolor{blue}{blue}) and our method augmented with
long short-term objective (\textcolor{green}{green}) in the hard setting
which has 8 agents and 3 campsites. The first layer of all mixing
networks in this experiment outputs a latent variable that only has
the size of $8$.}
\end{figure}

To highlight the role of the LSTRR regularisation, we intentionally
lower the performance of our method on the hard setting (8 agents
and 3 campsites) to roughly match that by QMIX by reducing the embedding
size in the first layer of the mixing network from $16$ (as used
in previous experiments) to $8$. However, augmenting our method with
the LSTRR regulariser greatly pushes the performance back, as shown
in Fig.~\ref{fig:figTransferHardSettingLSTR}.

\subsection{StarCraft Multi-Agent Challenge (SMAC) \label{subsec:StarCraft-Multi-Agent-Challenge}}

SMAC \citet{samvelyan19smac} is a recently benchmark for algorithms
for CTDE focusing on the StarCraft II Micro-Management in which each
unit is controlled by an agent.

\subsubsection{Implementation Details}

\begin{figure}
\begin{centering}
\includegraphics[width=1\columnwidth]{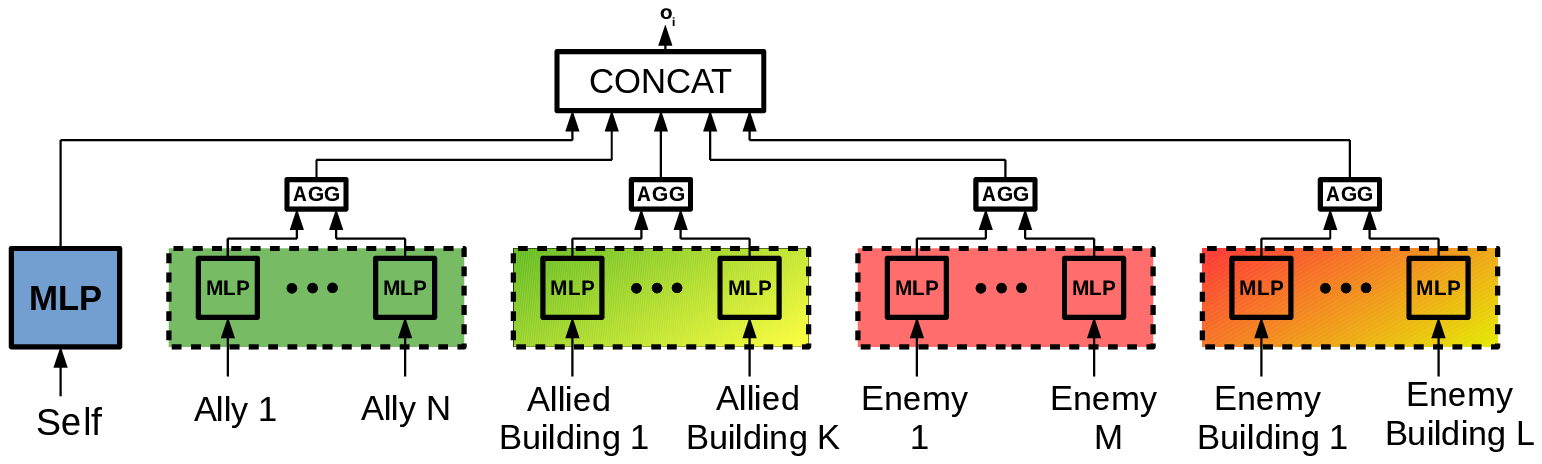}
\par\end{centering}
\caption{\label{fig:PreProcessObvs} Observation pre-processing architecture.}
\end{figure}

For each agent in SMAC, its observation is first pre-processed, as
shown in Fig.~\ref{fig:PreProcessObvs}. The purpose is to make the
observation $o_{i}$ of each agent independent of the number of agents
in a team by sharing weights between observations of objects of the
same types, e.g. allied troops, allied buildings, enemy troops, or
enemy buildings.

The strategy to choose actions during exploration (training phase)
is $\epsilon-$greedy. In the experiment, $\epsilon$ is annealing
from $0.15$ to $0.05$ during the first $50K$ time steps in the
source task and during the first $100K$ time steps in the target
task. Each agent does not observe itself ID to learn the index-free
policy. The batch size is $32$ episodes. The optimisation is RMSprop
with no momentum or weight decay, the learning rate is set as $5\times10^{-4}$,
and $\alpha=0.99$. We use the same method as proposed in SMAC paper
to evaluate our agents, which is after training for an interval of
$10K$ time steps, the learning team will decentralised execute. We
then measure the common rewards (for modified prey-predator) or the
test win rate (for SMAC).   

\subsubsection{Role-based Scenarios}

\begin{figure}[tp]
\begin{centering}
\includegraphics[width=1\columnwidth]{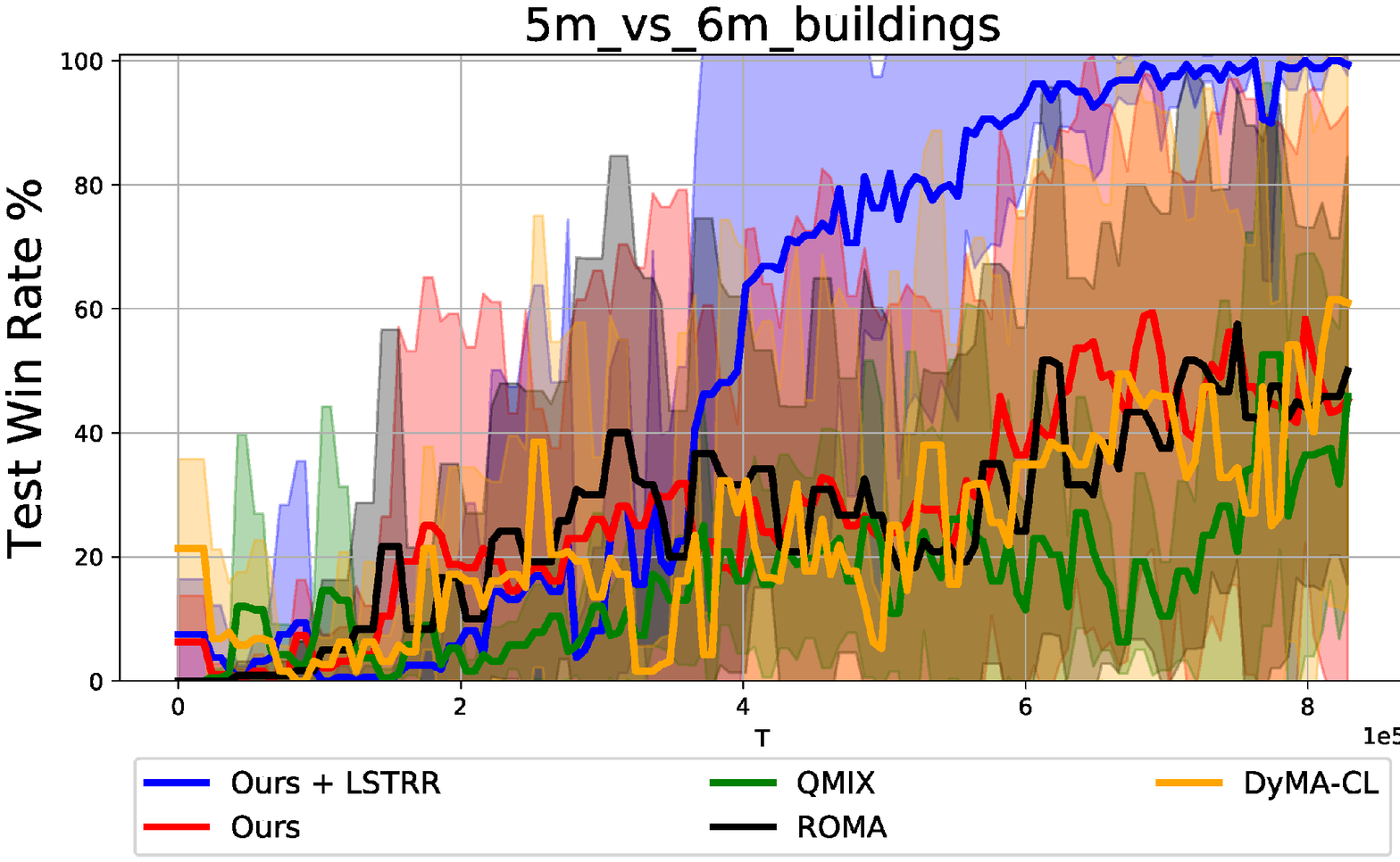}
\par\end{centering}
\caption{\label{figTestWonRate5m_vs_6m_buildings_smart_marine} {[}Best viewed
in colour{]} Test Win Rate (y-axis) vs. Training time steps (x-axis)
of teams trained by our method and baselines on the target task \emph{5m\_vs\_6m\_buildings}.}
\end{figure}
\begin{figure}[tp]
\begin{centering}
\includegraphics[width=1\columnwidth]{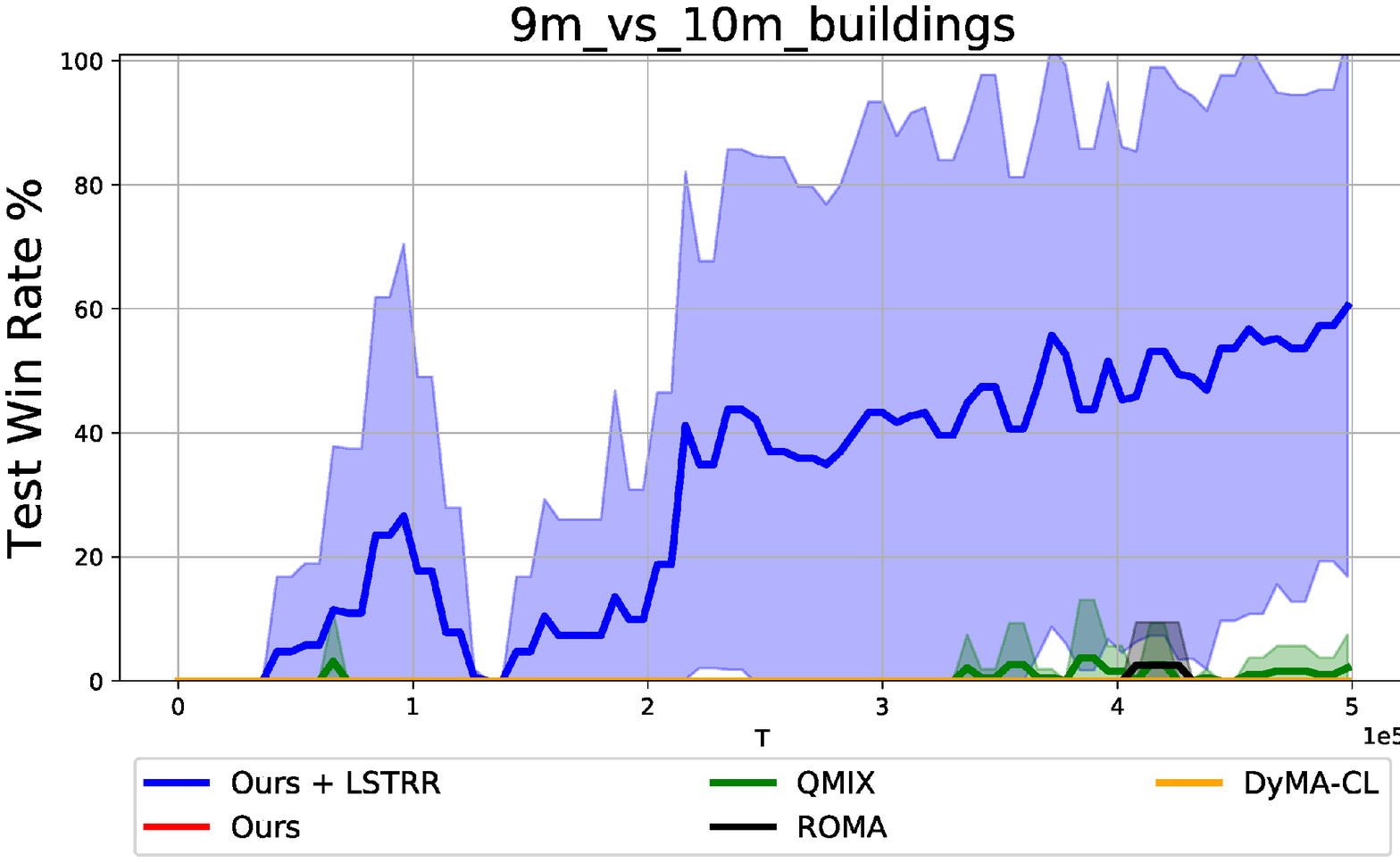}
\par\end{centering}
\caption{\label{figTestWonRate9m_vs_10m_buildings_smart_marine} Test Win Rate
(y-axis) vs. Training time steps (x-axis) of teams trained by our
method and baselines on the target task \emph{9m\_vs\_10m\_buildings}. }
\end{figure}

To show the ability of our architecture to transfer the individual
and mixer networks across team sizes, we construct a set of new scenarios
that require strict coordination in SMAC. More specifically, each
team has \emph{buildings}, which serve a similar purpose to the campsites
in our enriched prey-predator games in Section~\ref{subsec:Enriched-Prey-Predator-Games}\emph{.
}For example, in a particular setting called\emph{ 3m\_vs\_4m\_buildings},
the learned allied agents control three marines against four enemy
marines, and each team has one building to defend\emph{. }Similarly,
in \emph{5m\_vs\_6m\_buildings}, the learned allied agents control
five marines against six enemy marines; and each team has two buildings
to defend (see Fig.~\ref{fig:SMACBuildingsSmartMarine} in Section~\ref{sec:Introduction}
for an illustration). The scenario \emph{9m\_vs\_10m\_buildings} is
more difficult because there are two enemy marines always guarding
their buildings, and each team has three buildings. 

In our scenarios, the team needs not only to kill the opponent's troop
aggressively but also to defend its buildings and to destroy the enemy
buildings. This is because the game will be terminated if all buildings
of one team are destroyed. This forces agents to choose the defender
or attacker roles when they see fit. At the beginning of an episode,
there are enemy marines placed nearby the allied buildings; therefore,
defending own buildings is crucial to winning the game.

We compare our algorithm against three major baselines on SMAC: (1)
QMIX \citet{rashid2018qmix}: the mixer network captures non-linear
and monotonicity properties; (2) ROMA \citet{tonghan2020roma} which
learns emergent roles by hyper-network to generate weights of individual
agents; (3) DyMA-CL \citet{weixun2019few2more}: transfer individual
policy network. We consider the curriculum with the increasing team
size and difficulty. The DyMA-CL obtained good results before being
transferred to the bigger team size. Our individual and mixer networks
are first pre-trained on the team of size 3\emph{ }(\emph{3m\_vs\_4m\_buildings})\emph{,
}then transferred to train the team of size 5 (\emph{5m\_vs\_6m\_buildings}).
Finally, it is trained with the team of size 9 (\emph{9m\_vs\_10m\_buildings}).
Figs.~\ref{figTestWonRate5m_vs_6m_buildings_smart_marine} and \ref{figTestWonRate9m_vs_10m_buildings_smart_marine}
show that our networks trained with LSTRR regulariser can outperform
other baselines on the target tasks \emph{5m\_vs\_6m\_buildings }and\emph{
9m\_vs\_10m\_buildings}, respectively. Critically, without the LSTRR,
it is impossible to learn to play \emph{9m\_vs\_10m\_buildings} at
all (Fig.~\ref{figTestWonRate9m_vs_10m_buildings_smart_marine}). 

\subsubsection{Improving ROMA}

\begin{figure}
\begin{centering}
\subfloat[\label{figSMAC2s3z-1} 2s3z (Easy)]{\begin{centering}
\includegraphics[width=1\columnwidth]{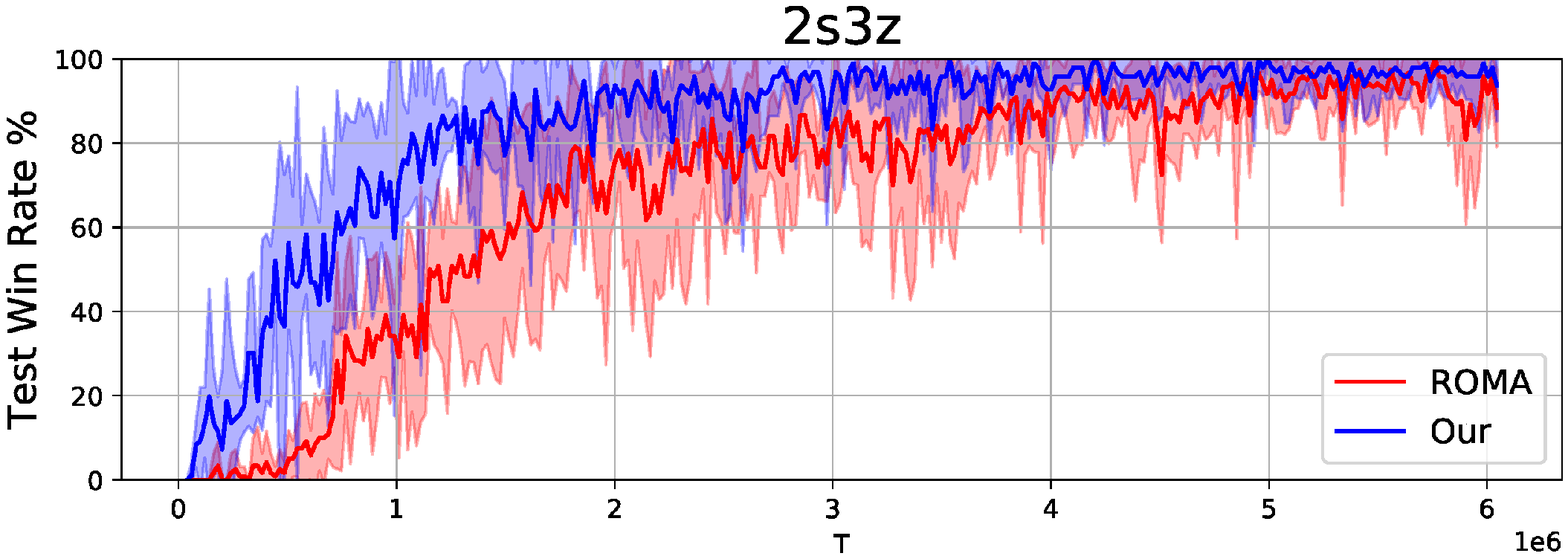}
\par\end{centering}
}
\par\end{centering}
\begin{centering}
\subfloat[\label{figSMACMMM2-1} MMM2 (Hard)]{\begin{centering}
\includegraphics[width=1\columnwidth]{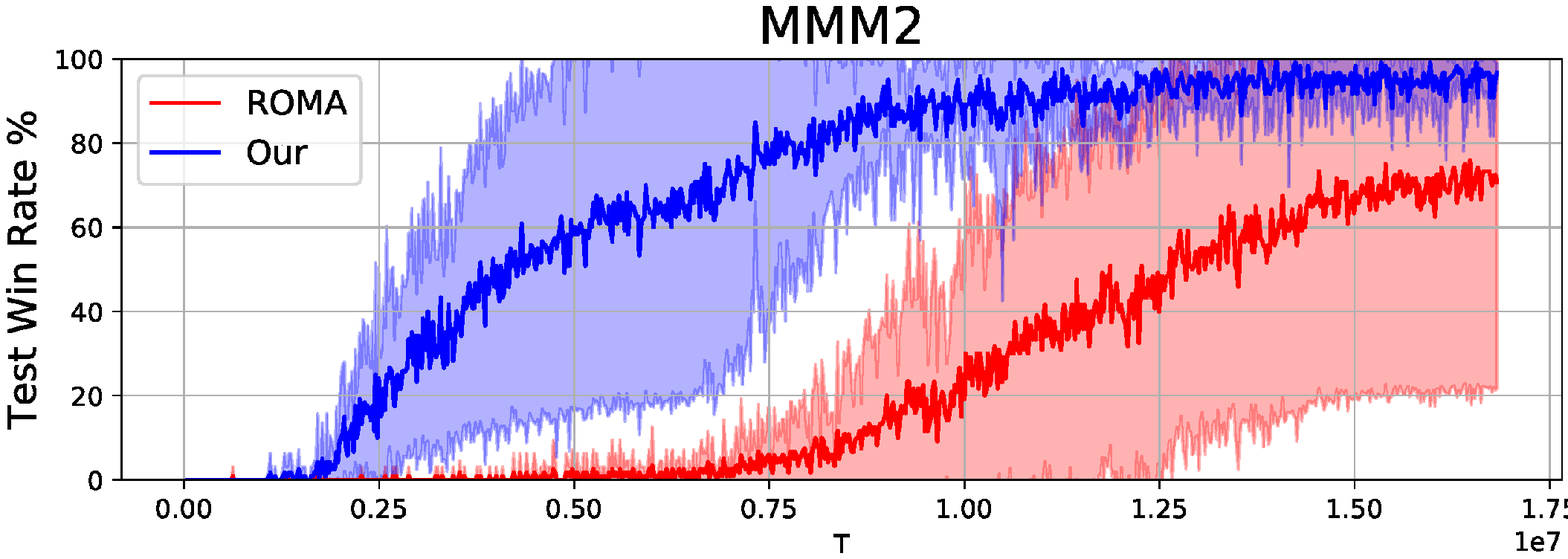}
\par\end{centering}
}
\par\end{centering}
\caption{\label{figSMAC-1} Test win rate vs. Training time steps in two SMAC
settings: (1) an easy setting and (2) a hard setting.}
\end{figure}

We conducted experiments to test our mixer with ROMA \citet{tonghan2020roma}
as individual policies (individual policies include a hyper-network
to generate roles) on two benchmark scenarios: (1) \emph{2s3z} (classified
as a Symmetric and Easy scenario): controlling 2 Stalkers and 3 Zealots
to defeat an enemy team which has the same units; (2) \emph{MMM2}
(classified as an Asymmetric and Hard scenario): controlling $1$
Medivac, $2$ Marauders and $7$ Marines to defeat an enemy team with
$1$ Medivac, $3$ Marauders and $8$ Marines. We incorporated our
architecture of the mixer into ROMA. Fig.~\ref{figSMAC-1} shows
that our architecture is more sample efficient than ROMA in both scenarios.
The team trained by ROMA has longer episode lengths compared to our
method (Fig.~\ref{figMMM2_eps_length}). To investigate this observation,
we compared the test battles of two methods. After $2,000,000$ training
time steps, even though teams trained by ROMA and ours could not learn
to defeat the enemy, there are significant differences in the agents'
behaviours. We observed the \emph{reward hacking} phenomenon in the
team trained by ROMA. In the middle of an episode, when some agents
were killed (the chance for the team to win is small), alive agents
retreated to the corner of the map (out of the sight of the enemy).
It is reasonable for individual agents to avoid being killed. However,
it induces wasteful samples for the training during the end phase
of the episodes. All agents in our team, in contrast, engage in the
battle and learn the optimal behaviours to win this game.

\begin{center}
\begin{figure}
\begin{centering}
\includegraphics[width=1\columnwidth]{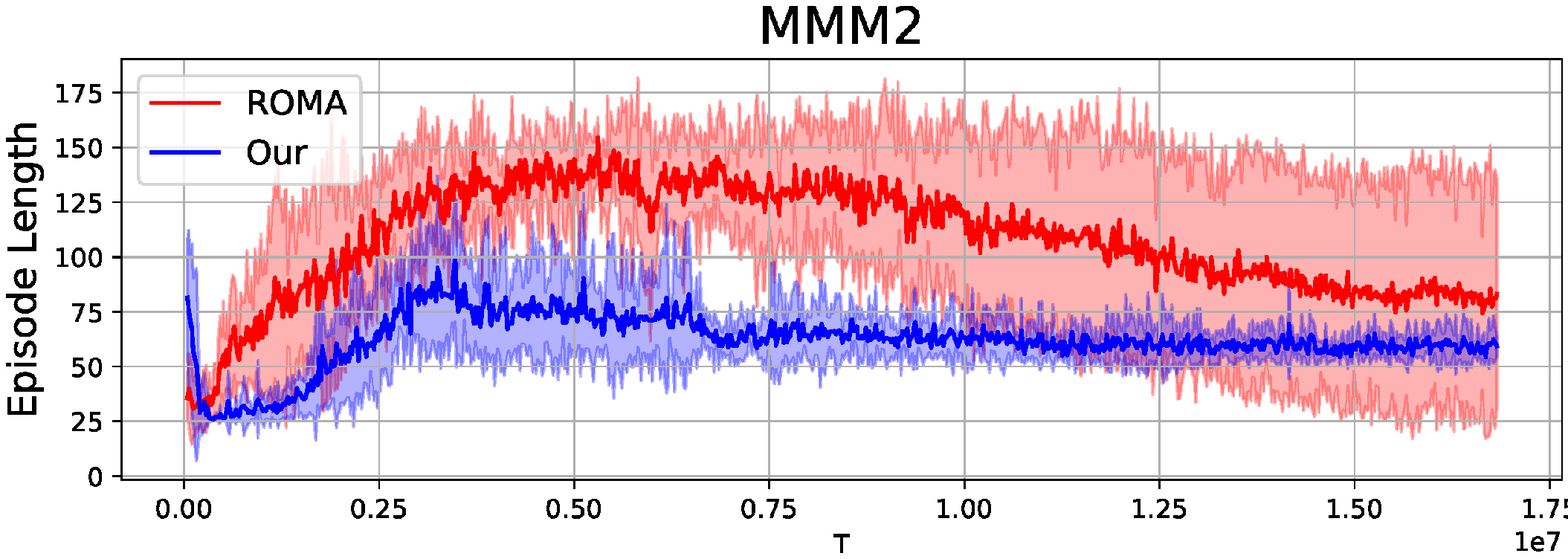}
\par\end{centering}
\caption{\label{figMMM2_eps_length} Episode length vs. Training time steps
in MMM2.}
\vspace{-5mm}
\end{figure}
\par\end{center}

\section{Related Works}

\paragraph{Team decomposition}

Value decomposition of a team reward in the CTDE paradigm was pioneered
by VDN \citet{sunehag2018value} which is a simple linear composition
of individual $Q$-values. Later, QMIX \citet{rashid2018qmix} improved
the composition function by bringing in the global state information
and relaxing the linearity into a monotonic linear composition. However,
this monotonicity restricts the class of value functions, especially,
it could fail to represent the optimal $Q^{*}$ \citet{rashid2020weighted,bohmer2020deep}.
To overcome this limitation, QTRAN \citet{son2019qtran} relaxed the
additivity and monotonicity by transforming all value functions to
satisfy the Individual-Global-Max (IGM) condition. Alternatively,
Qatten \citet{yang2020qatten} implemented multi-head attention to
generate the weights for agents based on their own properties. QPLEX
\citet{wang2020qplex}, on the other hand, used a duelling structure
for both joint and individual value functions, which can benefit
from off-line RL \citet{levine2020offline}. \vspace{-2mm}

\paragraph{Roles}

Another important line of work focuses on training agents to discover
behaviours and roles. ROMA \citet{tonghan2020roma} designed a role
embedding space and used a hyper-network to model the individual policies
conditioned on the role. The authors also introduced regularisers
based on diversity and identifiability to encourage the role emergence.
RODE \citet{wang2020rode} improved upon ROMA on role discovery by
decomposing the joint action space into regions associated with different
roles, thus learning a role selector and a role policy of lower temporal
resolution. In MARL, all agents need to coordinate their actions.
Individually exploring the environment could induce a large amount
of noisy rewards during training, a non-stationary learning phenomenon
\citet{chang2003all}. Different from ROMA which only motivates each
agent individually to explore optimal behaviour, MAVEN \citet{mahajan2019maven}
created a framework to explore the space of joint behaviours. While
both ROMA and MAVEN applied the technique of conditioning agent behaviours
on latent variables, Q-DPP \citet{yang2020multi} applied the determinantal
point process to improve the coordinated exploration when training
an RL team.  \vspace{-2mm}

\paragraph{Transfer learning in teams}

Training good individual policies for a small team size then adjusting
these policies for a large team size can be considered as an instance
of \emph{curriculum learning}. Recently, DyMA-CL \citet{weixun2019few2more}
proposed a training strategy on top of value based methods to transfer
across different team sizes. However, this method does not take into
account the learning and transferring roles of agents. Transferring
to new team sizes requires learning an index-free policy in which
the agent behaviour does not depend on its index in the team. In \citet{le2017coordinated},
authors proposed a method to learn roles from a set of experiences.
This method distinguishes roles by the trajectories induced by these
roles, while our method tries to learn roles based on their effects
at different time scales.

Learning multiple horizons has been empirically proved to improve
the performance of a single RL agent. The work in \citet{xu2018meta}
proposed to optimise the discount factor $\gamma$. The work in \citet{fedus2019hyperbolic}
suggested learning different $Q$-values for different discount factors
as auxiliary tasks. In \citet{romoff2019separating}, the value function
is broken down into different components based on smaller discount
factors. Recently, it has been suggested in \citet{amit2020discount}
the use of reduced discount factors to estimate the value function
in temporal difference learning, especially when the amount of data
is limited. However, these works only focus on using different discount
factors to facilitate training a single reinforcement learning agent;
we investigate the use of different discount factors in multi-agent
learning, realising under the concept of roles.

\section{Conclusion \label{sec:discuss}}

\inputencoding{latin9}We have introduced a new multi-agent reinforcement
learning framework to help scale an important paradigm known as centralised
training decentralised execution (CTDE). We redesigned the mixing
network in the popular QMIX framework to enable (i) learning with
arbitrary team sizes; (ii) assigning credits to roles, each of which
evaluates and attributes contributions from individual agents; and
(iii) curriculum learning process that starts from smaller teams and
progresses to large teams. We also contributed two suites of MARL
experiments to evaluate strongly cooperative CTDE tasks that demand
the notion of roles dynamically played by the agents. One suite enriches
the Prey-Predator games to include more types, roles and skills. The
other suite of experiments extends the StarCraft II Micro-Management
tasks. We demonstrated that the proposed framework leads to faster
convergence and the emergence of roles and can succeed in certain
large team settings.

 \balance

\bibliographystyle{ACM-Reference-Format}
\bibliography{main}

\end{document}